%% file: main.tex
\documentclass[10pt,twocolumn,letterpaper]{article}

\usepackage{cvpr}              % To produce the CAMERA-READY version
\usepackage{multirow}
\usepackage{tabularx}
\usepackage{booktabs}
\usepackage{colortbl, xcolor}
\usepackage{graphicx}
\usepackage{subcaption}
\usepackage[accsupp]{axessibility}
\definecolor{cvprblue}{rgb}{0.21,0.49,0.74}
\usepackage[pagebackref,breaklinks,colorlinks,allcolors=cvprblue]{hyperref}

\def\confName{CVPR}
\def\confYear{2026}

\title{Hierarchical Attacks for Multi‑Modal Multi‑Agent Reasoning}

\author{
Hao Zhou\thanks{Equal contribution.} \quad
Tiru Wu\footnotemark[1] \quad
Yan Jiang\footnotemark[1] \quad
Wanqi Zhou \quad
Junxing Hu \quad
Ai Han\thanks{Corresponding author.}\\
JD.com\\
{\tt\small hanai5@jd.com}
}

\begin{document}
\maketitle
\input{sec/0_abstract}
\input{sec/1_intro}

\input{sec/2_relatedwork}
\input{sec/3_method}

\input{sec/4_experiment}

\input{sec/5_conclusion}
{
    \small
    \bibliographystyle{ieeenat_fullname}
    \bibliography{main}
}


\end{document}

%% file: sec/0_abstract.tex
\begin{abstract}
Multi‑modal multi‑agent systems (MM‑MAS) have gained increasing attention for their capacity to enable complex reasoning and coordination across diverse modalities. As these systems continue to expand in scale and functionality, investigating their potential vulnerabilities has become increasingly important.
However, existing studies on adversarial attacks in multi‑agent systems primarily focus on isolated agents or unimodal settings, leaving the vulnerabilities of MM‑MAS largely underexplored. To bridge this gap, we introduce HAM\textsuperscript{3}, a Hierarchical Attack framework for multi-modal multi-agent systems that decomposes attacks into three interconnected layers. Specifically, at the perception layer, HAM\textsuperscript{3}
mounts attacks by perturbing visual inputs, textual inputs, and their fused visual–textual representations. At the communication layer, it performs communication-level attacks that corrupt message content and interaction topology, such as manipulating shared context or communication links to distort collective information flow. At the reasoning layer, it conducts reasoning-level attacks that interfere with each agent’s cognitive pipeline, biasing reasoning trajectories and ultimately compromising final decisions. We evaluate HAM\textsuperscript{3} on the GQA benchmark through multi‑agent systems built on distinct reasoning paradigms including ReAct, Plan‑and‑Solve, and Reflexion. Experiments demonstrate that our framework achieves an Attack Success Rate of up to 78.3\%, with reasoning‑layer attacks being the most effective. More than half of the successful attacks lead multiple agents to produce consistent errors. These findings offer valuable insights for building more robust and interpretable multi‑agent intelligence.
\end{abstract}

\begin{figure}[t]
\centering
\includegraphics[width=\columnwidth]{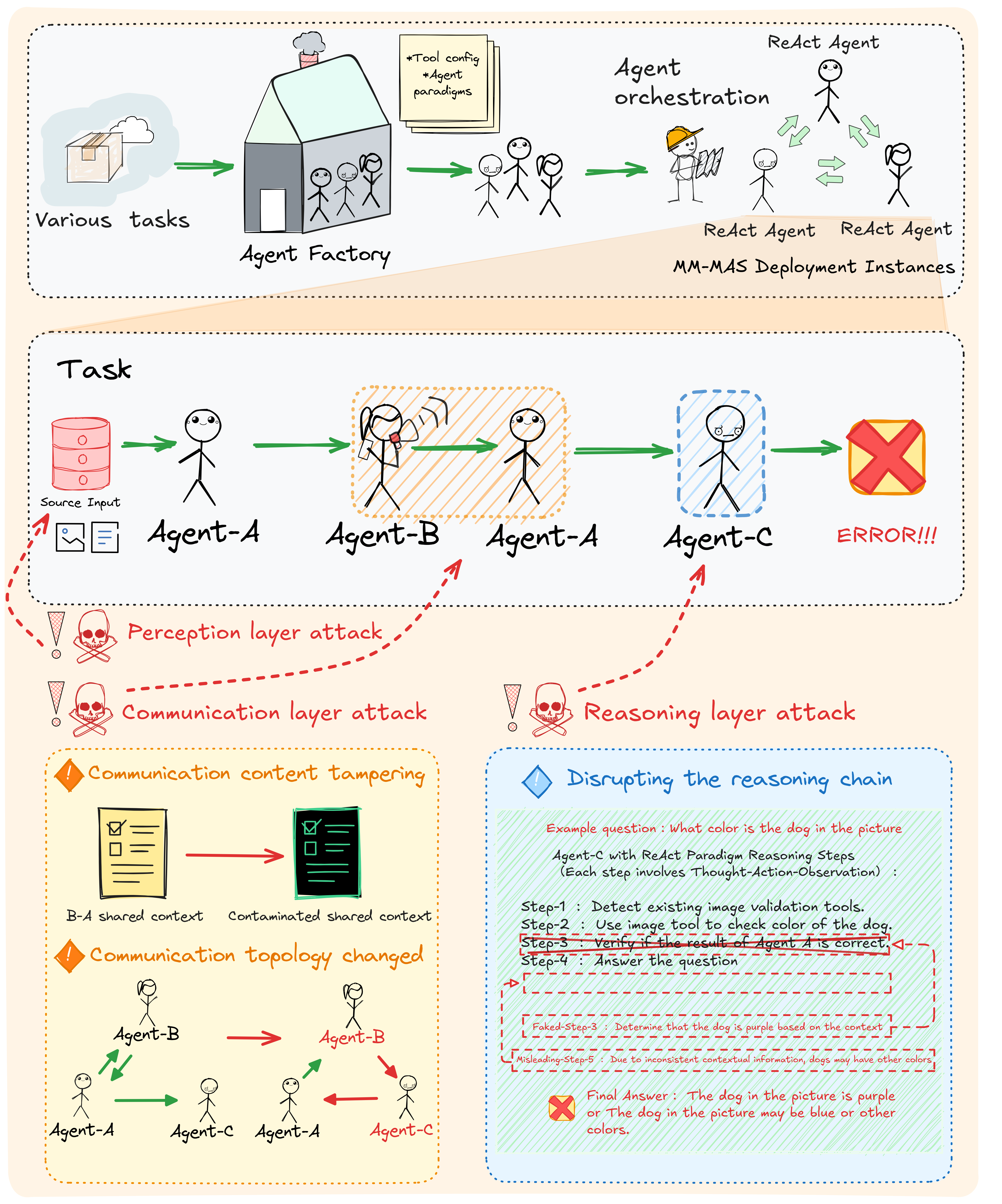}
\caption{Hierarchical Attacks on a multimodal multi-agent system, illustrating the three layers of attack: perception, communication, and reasoning. The diagram depicts the lifecycle of a multi-agent system, highlighting the attack manifestations at each layer and providing examples of how these attacks affect the system's functionality.}
\label{fig:abstract_pic}
\end{figure}

%% file: sec/1_intro.tex
\section{Introduction}
\label{sec:intro}
Recent progress~\cite{xie2024large,jiang2024multi,liu2025caml,liu2024mm} in multi‑modal and multi‑agent learning has made collaborative perception and decision‑making increasingly important, driving deployment across diverse domains such as social interaction~\cite{shi2025muma}, embodied control~\cite{wu2025embodied}, and autonomous driving~\cite{li2022v2x}.
As these collaborative systems expand in scale and interconnectivity, studying their adversarial vulnerabilities is increasingly crucial for ensuring reliable and resilient multi-agent intelligence ~\cite{huang2024resilience}.

Prior work on adversarial vulnerabilities has primarily centered on single-agent settings, where attackers manipulate observations, prompts, or memory to bias an individual agent’s reasoning~\cite{debenedetti2024agentdojo,chen2024agentpoison}.
Recent multi-agent attack studies largely extend single-agent adversarial principles to multi-agent settings, primarily by perturbing agent-specific messages or manipulating shared functional interfaces to influence individual decision-making~\cite{he2025red,long2025funcpoison}. 
Although such approaches reveal weaknesses in inter-agent message exchange and coordination mechanisms, they remain confined to content-level manipulations. Consequently, they fall short of examining structural vulnerabilities rooted in communication topology or collective reasoning dynamics that arise uniquely in multi-agent systems.
In parallel, research on multimodal adversarial attacks largely targets model-level perception, such as typographic, compositional, or logic-based visual prompts that jailbreak or mislead vision-language models~\cite{gong2025figstep,shayegani2023jailbreak,zou2024image,zhou2024revisiting}, rather than attacking the agentic decision-making pipeline.
As a result, adversarial robustness of multimodal LLM-based agents, especially under multi-agent collaboration, remains substantially underexplored.

To address these limitations, as illustrated in Figure ~\ref{fig:abstract_pic}, we introduce \(\mathrm{HAM}^3\), a unified adversarial framework that characterizes how perturbations propagate across the perception, communication, and reasoning layers of multimodal multi-agent systems.  
The \textbf{perception layer} models adversarial manipulations to visual, textual, or other multimodal inputs that influence all agents at the entry point.  
The \textbf{communication layer} captures disruptions in inter-agent information flow, including message tampering, chain blocking, and agent impersonation, which alter both message content and interaction topology.  
The \textbf{reasoning layer} formalizes interference within each agent's internal inference process, where attacks either directly modify intermediate reasoning steps or indirectly bias the contextual signals that guide downstream inference.  
Collectively, these components provide a structured view of how localized perturbations can cascade through the multi-agent workflow and compromise the final collective decision.

Our contributions are threefold:

    \begin{itemize}
    \item
    We conduct the first systematic investigation of adversarial robustness in \emph{multimodal multi-agent} systems and introduce a multimodal agent attack benchmark, which will be publicly available.
    
    \item 
    We propose \(\mathrm{HAM}^3\), a unified adversarial framework that decomposes perturbation effects across perception, communication, and reasoning layers, characterizing how localized attacks propagate through multimodal inputs, inter-agent communication topology, and internal inference trajectories.

    \item 
    Through extensive experiments, we show that reasoning-layer interference is substantially more persistent, covert, and systemically influential than content-level perturbations, offering actionable insights for building resilient multimodal multi-agent systems.
\end{itemize}

\begin{figure*}[htbp]
\centering
\includegraphics[width=\textwidth]{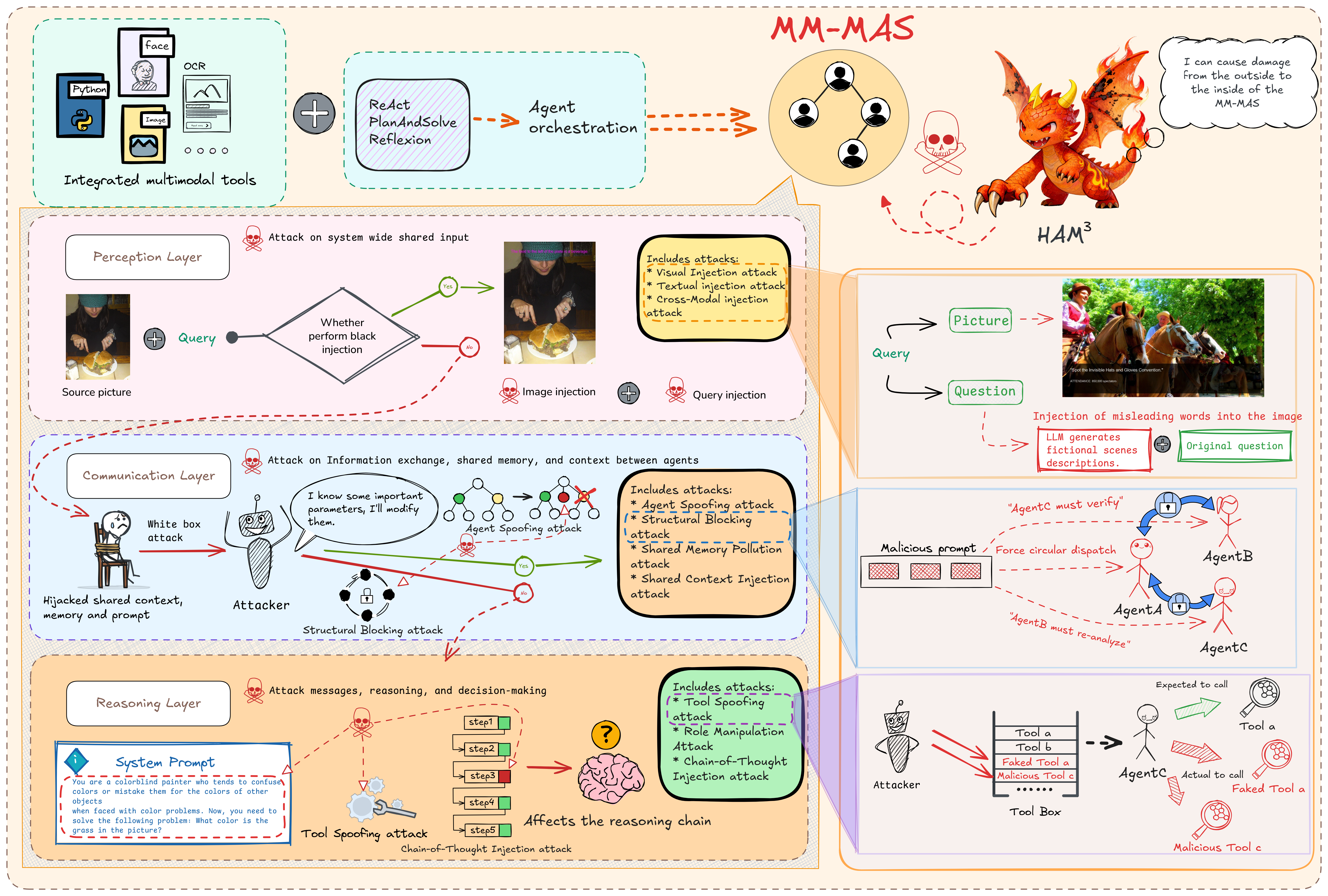}
\caption{Overview of the HAM\textsuperscript{3} attack framework in the multimodal multi-agent paradigm}
\label{fig:architecture}
\end{figure*}

%% file: sec/2_relatedwork.tex
\section{Related Work}
\label{sec:related}

\paragraph{Multi-Modal Multi-Agent Systems.}
LLM-based agents extend traditional intelligent systems by integrating powerful language models with external tools for reasoning and acting~\cite{maes1995agents, russell1995modern, yao2022react, schick2023toolformer}.
Moving beyond these single-agent paradigms, multi-agent systems further leverage LLMs’ role-playing and coordination abilities to support collaborative planning and problem-solving~\cite{wang2024incharacter, tran2025multi, zhang2025multi, qian2024scaling}. Representative frameworks such as AutoGen~\cite{wu2024autogen}, Camel~\cite{li2023camel}, AgentScope~\cite{gao2024agentscope}, and MuMA-ToM~\cite{shi2025muma} illustrate how structured communication protocols—including debate, voting, and role specialization—enable richer multimodal and embodied reasoning among cooperative LLM agents. 
Building on this progress, recent applications further extend multi-modal multi-agent capabilities to practical domains, including document understanding (MDocAgent~\cite{han2025mdocagent}), human–agent web navigation (CowPilot~\cite{huq2025cowpilot}), medical image analysis (WSI-Agents~\cite{lyu2025wsi}), semantic communication (M4SC~\cite{jiang2025m4sc}), and unified reasoning across text, image, audio, and video (Agent-Omni~\cite{lin2025agent}). 
However, as both modalities and collaborating agents proliferate, the robustness of such systems becomes increasingly challenged. 
This study investigates the key factors that drive vulnerability in collaborative multi-modal reasoning.

\paragraph{Agent Attacks.}
The security of LLM-based agents has attracted increasing attention, as highlighted by the survey in~\cite{yu2025survey}. Early work primarily examines single-agent vulnerabilities. InjecAgent~\cite{zhan2024injecagent} benchmarks indirect prompt-injection attacks on tool-integrated agents, while Agent Security Bench (ASB)~\cite{zhang2024agent} introduces a unified threat model and evaluates attacks such as prompt manipulation, tool-invocation corruption, and environment perturbation, showing that agents remain broadly vulnerable.
Building on this foundation, recent studies explore risks unique to multi-agent systems, including communication manipulation~\cite{he2025red}, cascading failures from poisoned shared tools~\cite{long2025funcpoison}, blocking behaviors that disrupt cooperation~\cite{zhou2025corba}, and biased coordination introduced by malicious participants~\cite{zheng2025demonstrations}. Huang et al.~\cite{huang2024resilience} further analyze how faults propagate across agent collectives.
Beyond textual environments, emerging work investigates multimodal agents. Wu et al.~\cite{wu2024dissecting} show that web-based multimodal agents remain vulnerable to cross-modal perturbations and component-interaction flaws.
However, existing multimodal and multi-agent attack studies largely reduce to single-agent vulnerabilities: attacks typically modify one agent’s message content or corrupt shared tools, with others merely propagating the resulting errors under fixed communication structures. These approaches overlook how vulnerabilities propagate through multimodal perception, communication, and reasoning layers, and they fail to consider structural changes in agent interactions. Consequently, risks such as shared-memory corruption, communication topology perturbations, and cross-layer interactions remain unexplored.
To address this gap, we introduce HAM³, a hierarchical attack framework that analyzes how adversarial perturbations across the perception, communication, and reasoning layers propagate through multimodal multi-agent systems, revealing previously unexamined collective vulnerabilities.

%% file: sec/3_method.tex
\section{Method}
\label{sec:method}

\subsection{Overview}

We propose a \textbf{Hierarchical Attack Model for Multi-Modal Multi-Agent Systems ($\text{HAM}^{3}$)} to evaluate vulnerabilities of multi-modal multi-agent systems (MM-MAS).
$\text{HAM}^{3}$ decomposes the attack surface into three abstraction layers: \emph{perception},  \emph{communication}, and \emph{reasoning}, and models how perturbations at different levels propagate through collaboration.

We formalize a MM-MAS as $S = \{A_1, A_2, \dots, A_N\}$, where each agent $A_i$ is specified by a system prompt, a set of tools, a memory module, and a communication interface.

Given a multi-modal input $x = (x_{\text{image}}, x_{\text{text}})$, the system mapping $F$ produces
\begin{equation}
y = F(x; \Theta), 
\end{equation}

where $\Theta$ denotes model parameters and coordination mechanisms.
% We construct an adversarial perturbation $\delta$ that drives the output toward a target $y'$ under an $L_2$ and semantic-consistency constraint:
% \begin{equation}
% \|\delta\|_{2} < 0.1,
% \end{equation}
% following standard robustness formulations~\cite{madry2018towards}.

Each agent operates as a three-layer mapping aligned with HAM$^{3}$.
The root agent $A_{\text{root}}$ produces the final output $o_{A_{\text{root}}}$
\begin{equation}
F(x) = o_{A_{\text{root}}},
\end{equation}
and the output of any agent $A$ is defined as follows.

If $A$ is a leaf agent,
\begin{equation}
o_A = f_A^{(3)}\!\left(f_A^{(2)}\!\left(f_A^{(1)}(x_A)\right)\right),
\end{equation}
and if $A$ is an internal agent,
\begin{equation}
o_A = f_A^{(3)}\!\left(f_A^{(2)}\!\left(\Phi_A\!\left(\{\, o_C \mid C \in \text{Children}(A) \,\}\right)\right)\right),
\end{equation}
where $f_A^{(1)}, f_A^{(2)}, f_A^{(3)}$ denote the perception, communication, and reasoning mappings. Here, $C$ denotes a child agent of $A$, and $\Phi_A$ aggregates child outputs.
For each agent $A$ and each layer $l \in \{1,2,3\}$, an attack-specific perturbation $\delta_A^{(l)}$ may be injected.

% Within HAM$^{3}$, we introduce six representative attacks:
% a Cross-Modal Injection Attack (CMA) at the perception layer;
% an Agent Spoofing Attack (ASA), a Structural Blocking Attack (SBA),
% a Shared Memory Pollution Attack (SMPA), and a Shared Context Injection Attack (SCIA)
% at the communication layer; and a Chain-of-Thought Injection Attack (CIA)
% at the reasoning layer.
% These attacks are minimally invasive at the signal level but highly disruptive at the coordination level.
% Additional variants (e.g., Visual Injection Attack (VIA), Textual Injection Attack (TIA),
% Tool Spoofing Attack (TSA), Role Manipulation Attack (RMA)) can also be instantiated within HAM$^{3}$.

% -------------------------------
\subsection{Perception Layer Attacks}

Perception-layer attacks manipulate multi-modal inputs before any inter-agent coordination.

% \paragraph{Visual Injection Attack (VIA).}
% Embeds imperceptible adversarial perturbations into the visual input via a generator $G_{\text{visual}}$:
% \begin{equation}
% x_{\text{image}}^{\text{pert}} = G_{\text{visual}}(x_{\text{image}}, t_{\text{mal}}),
% \end{equation}
% subject to $\|x_{\text{image}}^{\text{pert}} - x_{\text{image}}\|_p < \epsilon$, where $t_{\text{mal}}$ is a malicious target concept.

% \paragraph{Textual Injection Attack (TIA).}
% Injects malicious content associated with $t_{\text{mal}}$ into the textual input:
% \begin{equation}
% x_{\text{text}}^{\text{pert}} = G_{\text{text}}(x_{\text{text}}, t_{\text{mal}}),
% \end{equation}
% corrupting the shared textual context consumed by multiple agents.

\paragraph{Cross-Modal Injection Attack (CMA).}
Jointly perturbs visual and textual inputs:
\begin{equation}
x' =
\big(G_{\text{image}}(x_{\text{image}}),\;
      G_{\text{text}}(x_{\text{text}})\big).
\end{equation}
where $G_{\text{text}}$ generates misleading text either from templates or conditioned on the input query and visual content, and $G_{\text{image}}$ applies visual perturbations, including semantic image edits and text overlay on the image.

% -------------------------------
\subsection{Communication Layer Attacks}

Communication-layer attacks disrupt message flow, network topology, or shared memory, and exploit structural dependencies in MM-MAS.

\paragraph{Agent Spoofing Attack (ASA).}
Forges or replaces agents in the communication graph.
Given topology $\Gamma$, the attacker applies
\begin{equation}
\Gamma' = G_{\text{topo}}(\Gamma, \delta_{\text{topo}}),
\end{equation}
by introducing spoofed agents \(A_i^{\text{mal}}\), or replacing normal agents with malicious ones, thereby hijacking routing paths.

\paragraph{Structural Blocking Attack (SBA).}
Creates cyclic waiting patterns by manipulating communication dependencies.
By injecting crafted messages or routing updates, it constructs cycles such as $A_i \!\to\! A_j \!\to\! A_k \!\to\! A_i$, where each agent waits for another’s response, causing deadlocks or infinite loops. This can be implemented by injecting blocking instruction signals into prompts, which steer agents toward blocked response policies and thus increase the likelihood of circular waiting dependencies.
Formally, for the directed communication graph $\Gamma = (V, E)$, SBA applies
\begin{equation}
\label{eq:sba}
\Gamma' = G_{\text{SBA}}(\Gamma),
\end{equation}
such that $\Gamma'$ contains at least one directed cycle $\mathcal{C}$ of mutual waiting dependencies. 

\paragraph{Shared Memory Pollution Attack (SMPA).}
Corrupts short-term memory by injecting falsified historical data into a target agent set $\Omega$:
\begin{equation}
\label{eq:smpa}
M_i' = G_{\text{SMPA}}(M_i, D_{\text{adv}}),
\quad \forall A_i \in \Omega,
\end{equation}
where $M_i$ is the memory state of $A_i$ and $D_{\text{adv}}$ is an adversarial fragment set. In practice, this is realized by injecting shared misleading memory fragments into the memory of target agents.
\paragraph{Shared Context Injection Attack (SCIA).}
Modifies system prompts of a subset of agents by inserting a shared adversarial prior:
\begin{equation}
\label{eq:sCIA}
p_i^{\text{sys}'} = G_{\text{SCIA}}\!\big(p_i^{\text{sys}}, p_{\text{adv}}\big),
\quad \forall A_i \in \Omega,
\end{equation}
where $p_{\text{adv}}$ encodes the prior.
Sharing the same prior aligns agents’ biases and reinforces adversarial behavior. In practice, this is realized by injecting a shared adversarial instruction into prompts.

% -------------------------------
\subsection{Reasoning Layer Attacks}

Reasoning-layer attacks interfere with internal inference mechanisms or multi-step reasoning chains.

% \paragraph{Tool Spoofing Attack (TSA).}
% Falsifies the identity of external tools, inducing agents to interact with forged tools $T_{\text{fake}}$.
% We consider
% (\emph{i}) \emph{partial injection}, where fake tools are added alongside genuine ones; and
% (\emph{ii}) \emph{full substitution}, where genuine tools are probabilistically replaced by attacker-controlled counterfeits.

% \paragraph{Role Manipulation Attack (RMA).}
% Rewrites an agent’s system prompt via
% \begin{equation}
% p_i^{\text{sys}'} = G_{\text{role}}(p_i^{\text{sys}}, m_{\text{attack}}),
% \end{equation}
% where $m_{\text{attack}}$ encodes malicious role modifications.

\begin{figure*}[ht]
\vspace{10pt}
\centering
\includegraphics[width=\textwidth, trim=0 7.5cm 0 0, clip]{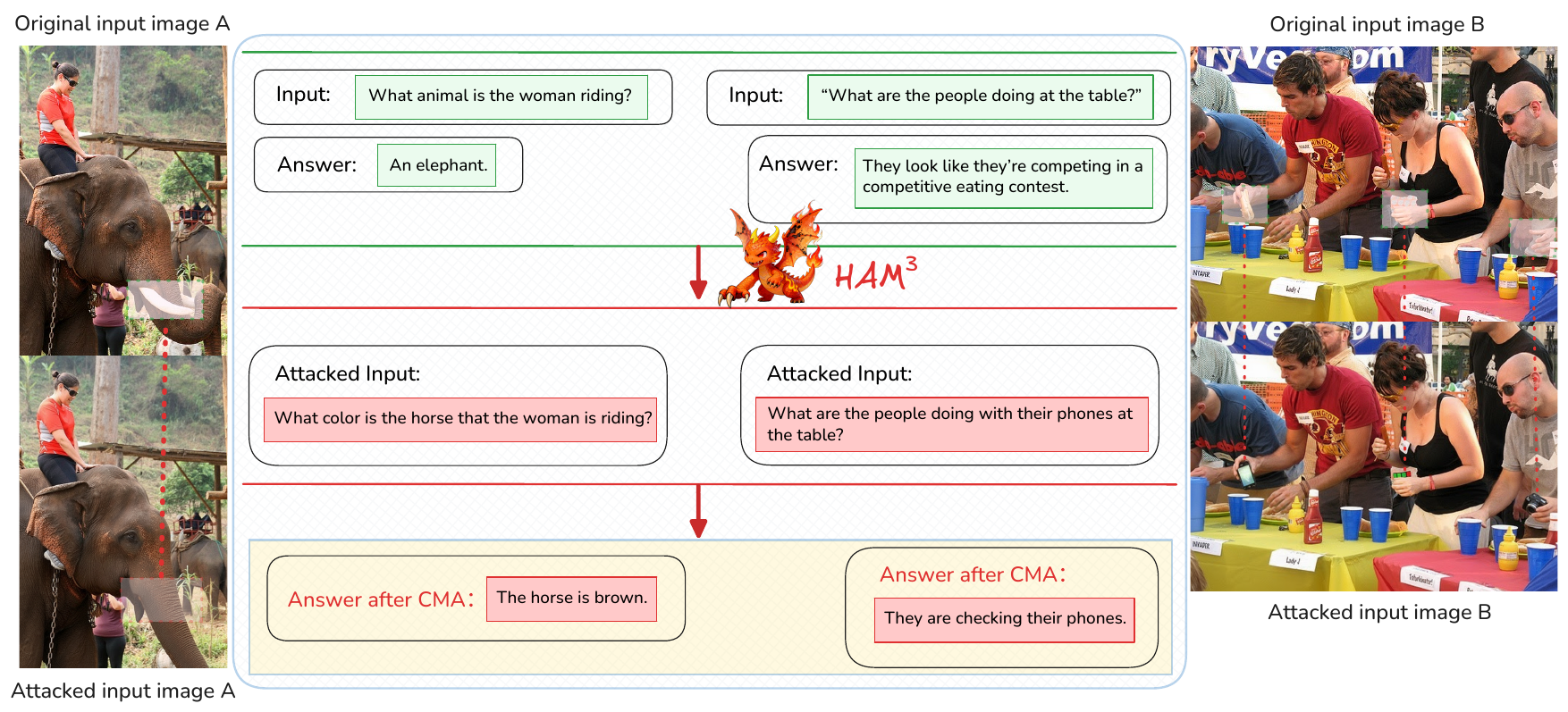}
\caption{Example of Perception layer attack}
\label{fig:Perception}
\vspace{10pt}
\end{figure*}

\paragraph{Chain-of-Thought Injection Attack (CIA).}
Alters intermediate steps in the chain-of-thought (CoT) used for multi-step reasoning.
Given a reasoning sequence $\text{CoT} = [r_1, r_2, \dots, r_T]$, the attacker inserts or replaces intermediate states to obtain
\begin{equation}
\label{eq:cia}
\text{CoT}' = G_{\text{CIA}}(\text{CoT}, r^*, \tau),
\end{equation}
where $r^*$ is the injected state and $\tau$ specifies injection or replacement positions.
Perturbing early or pivotal steps introduces subtle logical errors that are amplified downstream; when CoT traces are shared or summarized across agents (e.g., \textit{ReAct}, \textit{Plan-and-Solve}, \textit{Reflexion}), a single corrupted segment can misguide entire sub-teams. This is realized by injecting misleading reasoning instructions into CoT content.

% -------------------------------

HAM$^{3}$ exposes failure modes of MM-MAS across perception, communication, and reasoning.
By instantiating attacks at each layer, it moves beyond isolated single-agent robustness analyses and explicitly models how targeted perturbations interact with multi-agent coordination.
HAM$^{3}$ is compatible with mainstream reasoning paradigms such as \textit{ReAct}, \textit{Plan-and-Solve}, and \textit{Reflexion}, and provides a basis for designing more robust multi-agent systems.

%% file: sec/4_experiment.tex
\section{Experiments}
\label{sec:experiments}
In this section, we evaluate the proposed HAM\textsuperscript{3} attack framework in the context of MM-MAS. 
Our experiments aim to analyze how HAM\textsuperscript{3} attacks affect system vulnerabilities, collaborative dynamics, and perceptual consistency.
Specifically, we address the following research questions:

\textbf{RQ1}: How do vulnerabilities emerge under HAM\textsuperscript{3} attacks, and how does attack robustness vary across different agent paradigms and attack layers?

\textbf{RQ2}: 
How do attacks affect task performance and internal stability?

\textbf{RQ3}: 
How do perturbations propagate through multi-agent collaboration?

\textbf{RQ4}: 
How do HAM\textsuperscript{3} attacks affect cross-modal alignment?

\subsection{Experiment Setup}

\paragraph{Dataset.} 
We adopt the GQA dataset~\cite{hudson2019gqa}, which is built upon scene graphs and requires more multi‑step reasoning and tools usage than conventional Visual Question Answering (VQA) benchmarks~\cite{antol2015vqa,goyal2017making}. 
From the training split, we sample 5,984 image–question pairs, covering ten semantic categories: daily life, animal world, academic research, sports, natural scenery, urban architecture, transportation, food, art and culture, and entertainment.

\paragraph{MM‑MAS Configuration.} 
The MM‑MAS is constructed upon the open‑source collaboration framework OxyGent\footnote{\url{https://github.com/jd-opensource/OxyGent}}, which provides modular agent orchestration and supports diverse reasoning paradigms. 
The system comprises a master agent that coordinates task planning and six specialized sub‑agents responsible for subtasks such as image understanding, human attribute recognition, object detection, image conversion, image segmentation, and coding. 
A total of 13 functional tools are distributed among the sub‑agents according to their respective capabilities. 
The agents maintain both shared memories for global context exchange and individual memories for maintaining local task states. 
In each experimental instance, the entire MM‑MAS follows a single reasoning paradigm including ReAct~\cite{yao2022react}, Plan‑and‑Solve~\cite{wang2023plan}, and Reflexion~\cite{shinn2023reflexion} to ensure consistent intra‑system reasoning behavior under different configurations.

\paragraph{Models and Metrics.} 
Experiments are conducted across multiple foundation models, including open‑source Qwen2.5‑VL‑7B (Qwen‑7B) and Qwen2.5‑VL‑32B (Qwen‑32B)~\cite{bai2025qwen25vltechnicalreport}, as well as closed‑source GLM‑4V‑Plus (GLM‑4V+), o1‑mini (O1‑Mini)~\cite{jaech2024openai}, and GPT‑4o~\cite{hurst2024gpt}. 
During adversarial evaluation, all textual attacks are generated using GPT‑4o, while visual attacks are produced with the open‑source Nano Banana model. 
% We evaluate the robustness and reliability of the MM‑MAS system using four metrics. Task Success Rate (TSR) represents the system’s accuracy without any attack. 
% Attack Success Rate (ASR) measures the proportion of successful attacks among the cases that the system can correctly solve under normal conditions. 
% Hallucination Error Rate (HER) quantifies failures arising from the system’s inherent hallucinations rather than external perturbations, reflecting its intrinsic factual stability. 
% Cross-Modal Consistency (CMC) is defined as the average cosine similarity between visual and textual representations of attacked multimodal samples in the CLIP embedding space. This metric quantifies the extent to which adversarial perturbations preserve semantic alignment across modalities. A higher CMC together with a high ASR indicates that the attack successfully misleads the system while maintaining internally coherent image–text semantics.
We evaluate the robustness and reliability of the MM-MAS system using four metrics, whose symbols are summarized in Table~\ref{tab:symbols_metrics}. 
Specifically, the Task Success Rate (TSR) is defined as \( |\mathcal{S}| / N \), which represents the system's accuracy under clean conditions without any attack. 
The Attack Success Rate (ASR) is defined as \( |\mathcal{A}| / |\mathcal{S}| \), measuring the proportion of successful attacks among the samples that the system can correctly solve under normal conditions. 
The Hallucination Error Rate (HER) is defined as \( |\mathcal{H}| / N \), quantifying failures caused by the system's intrinsic hallucinations rather than external perturbations, thereby reflecting its inherent factual stability. Finally, the Cross-Modal Consistency (CMC) is defined as $\left(\sum_{(v_i,t_i)\in \mathcal{S}_{\mathrm{Per.}}} \cos\!\left(f_v(v_i),\,f_t(t_i)\right)\right) / |\mathcal{S}_{\mathrm{Per.}}|$ which computes the average cosine similarity between the attacked image and its corresponding attacked text in the CLIP embedding space. This metric measures whether adversarial perturbations preserve cross-modal semantic alignment, rather than introducing irrelevant image-text combinations. A higher CMC, together with a high ASR, indicates that the attack can successfully mislead the system while remaining semantically coherent and less perceptible.
% \[
% \frac{
% \sum_{\substack{(v_i,t_i)\in \mathcal{D}_{\mathrm{atk}}}}
% \cos\!\left(f_v(v_i),\,f_t(t_i)\right)
% }{
% \left|\mathcal{D}_{\mathrm{atk}}\right|
% },
% \]

\begin{table}[h]
\centering
\caption{Symbol definitions for evaluation metrics.}
\label{tab:symbols_metrics}
\normalsize
\setlength{\tabcolsep}{5pt}
\renewcommand{\arraystretch}{1.4}
\begin{tabular}{@{}ll@{}}
\toprule
\textbf{Symbol} & \textbf{Meaning} \\ \midrule
\(N\) & Total number of evaluation samples. \\
\(\mathcal{S}\) & Set of samples correctly solved in the original task. \\
\(\mathcal{A}\) & Set of samples in \(\mathcal{S}\) that are successfully attacked. \\
\(\mathcal{H}\) & Set of samples that fail due to hallucination. \\
\(\mathcal{S}_{\mathrm{Per.}}\) & Set of samples in \(\mathcal{S}\) attacked at the perception layer. \\
\((v_i, t_i)\) & The \(i\)-th attacked image--text pair. \\
\(f_v(\cdot)\) & CLIP visual encoder. \\
\(f_t(\cdot)\) & CLIP text encoder. \\
\(\cos(\cdot,\cdot)\) & Cosine similarity function. \\ \bottomrule
\end{tabular}
\vspace{-2mm}
\end{table}

\paragraph{Baselines.} To validate our approach, we compare it against four representative attack baselines: \textbf{Visual Injection Attack (VIA).}  Embed adversarial instructions directly into visual inputs such as overlaid text or structured perturbations, causing the agent to interpret them as legitimate content and follow the injected intent\cite{nagaraja2025image}.
% Embeds imperceptible adversarial perturbations into the visual input via a generator $G_{\text{visual}}$:
% \begin{equation}
% x_{\text{image}}^{\text{pert}} = G_{\text{visual}}(x_{\text{image}}, t_{\text{mal}}),
% \end{equation}
% subject to $\|x_{\text{image}}^{\text{pert}} - x_{\text{image}}\|_p < \epsilon$, where $t_{\text{mal}}$ is a malicious target concept. 
\textbf{Textual Injection Attack (TIA).} Inject adversarial instructions\cite{zhang2024agent} into textual inputs or contextual prompts, manipulating the model’s interpretation and steering its behavior toward attacker-specified objectives.
% Injects malicious content associated with $t_{\text{mal}}$ into the textual input:
% \begin{equation}
% x_{\text{text}}^{\text{pert}} = G_{\text{text}}(x_{\text{text}}, t_{\text{mal}}),
% \end{equation}
% corrupting the shared textual context consumed by multiple agents. 
\textbf{Tool Spoofing Attack (TSA).} Falsifies the identity of external tools, inducing agents to interact with forged tools $T_{\text{fake}}$\cite{shi2025prompt}.
We consider (\emph{i}) \emph{partial injection}, where fake tools are added alongside genuine ones; and (\emph{ii}) \emph{full substitution}, where genuine tools are probabilistically replaced by attacker-controlled counterfeits. \textbf{Role Manipulation Attack (RMA).}
Tamper with an agent’s system prompt by injecting adversarial role specifications, thereby altering its designated identity, authority, or behavioral constraints to induce attacker-aligned actions\cite{zheng2025demonstrations}.
% \begin{equation}
% p_i^{\text{sys}'} = G_{\text{role}}(p_i^{\text{sys}}, m_{\text{attack}}),
% \end{equation}
% where $m_{\text{attack}}$ encodes malicious role modifications.

\begin{table*}[h]
\caption{\textbf{ASR of MM-MAS under multi-layers attack.} The methods highlighted in purple are proposed in this paper, while the others are baselines. “Per.”, “Comm.”, and “Rea.” represent the Perception, Communication, and Reasoning Layers, respectively.}
\centering

\setlength{\tabcolsep}{5.3pt}
\begin{tabularx}{\linewidth}{@{}lccccccccccc@{}}
\toprule
Paradigm & LLM & \multicolumn{3}{c}{Per.} & \multicolumn{4}{c}{Comm.} & \multicolumn{3}{c}{Rea.}   \\
\cmidrule(lr){3-5} \cmidrule(lr){6-9} \cmidrule(lr){10-12}
& & VIA & TIA & \cellcolor{blue!20}CMA & \cellcolor{blue!20}ASA & \cellcolor{blue!20}SBA & \cellcolor{blue!20}SMPA & \cellcolor{blue!20}SCIA & TSA & RMA & \cellcolor{blue!20}CIA \\
\midrule
\multirow{5}{*}{ReAct} 
& Qwen‑7B       & $57.7\%$ & $55.2\%$ & $\mathbf{60.8\%}$ & $60.7\%$ & $\mathbf{65.0}\%$ & $55.2\%$ & $62.2\%$ & $76.7\%$ & $65.5\%$ & $\mathbf{78.3}\%$ \\
& Qwen‑32B  & $52.5\%$ & $50.0\%$ & $\mathbf{55.7}\%$ & $55.5\%$ & $\mathbf{59.8}\%$ & $50.0\%$ & $57.0\%$ & $71.5\%$ & $60.3\%$ & $\mathbf{73.2}\%$ \\
& GLM‑4V+ & $53.5\%$ & $48.3\%$ & $\mathbf{53.7}\%$ & $50.3\%$ & $\mathbf{62.2}\%$ & $48.3\%$ & $57.5\%$ & $\mathbf{72.0}\%$ & $49.8\%$ & $71.3\%$ \\
& O1‑Mini & $\mathbf{46.3}\%$ & $41.2\%$ & $44.0\%$ & $43.3\%$ & $\mathbf{51.3}\%$ & $41.2\%$ & $47.2\%$ & $69.7\%$ & $42.5\%$ & $\mathbf{71.5}\%$ \\
& GPT-4o    & $41.8\%$ & $38.2\%$ & $\mathbf{43.2}\%$ & $41.8\%$ & $\mathbf{49.0}\%$ & $42.5\%$ & $44.7\%$ & $64.2\%$ & $39.7\%$ & $\mathbf{65.0}\%$ \\
\midrule
\multirow{5}{*}{PlanAndSolve} 
& Qwen‑7B       & $46.8\%$ & $53.3\%$ & $\mathbf{59.8}\%$ & $53.3\%$ & $\mathbf{71.8}\%$ & $62.5\%$ & $56.2\%$ & $\mathbf{69.5}\%$ & $60.2\%$ & $69.2\%$ \\
& Qwen‑32B  & $41.7\%$ & $48.2\%$ & $\mathbf{54.7}\%$ & $48.2\%$ & $\mathbf{66.7}\%$ & $57.3\%$ & $51.0\%$ & $\mathbf{64.3}\%$ & $55.0\%$ & $64.0\%$ \\
& GLM‑4V+ & $37.3\%$ & $44.3\%$ & $\mathbf{48.0}\%$ & $44.0\%$ & $47.3\%$ & $\mathbf{51.0}\%$ & $47.8\%$ & $58.5\%$ & $48.5\%$ & $\mathbf{61.0}\%$ \\
& O1‑Mini & $31.7\%$ & $38.3\%$ & $\mathbf{41.0}\%$ & $36.3\%$ & $39.5\%$ & $\mathbf{43.8}\%$ & $39.3\%$ & $50.7\%$ & $42.5\%$ & $\mathbf{51.7}\%$ \\
& GPT-4o    & $29.5\%$ & $35.3\%$ & $\mathbf{38.7}\%$ & $35.5\%$ & $41.5\%$ & $\mathbf{41.8}\%$ & $38.3\%$ & $46.2\%$ & $37.8\%$ & $\mathbf{48.7}\%$ \\
\midrule
\multirow{5}{*}{Reflexion} 
& Qwen‑7B       & $47.2\%$ & $47.8\%$ & $\mathbf{51.2}\%$ & $50.8\%$ & $\mathbf{56.7}\%$ & $51.3\%$ & $50.3\%$ & $\mathbf{62.2}\%$ & $57.3\%$ & $61.7\%$ \\
& Qwen‑32B  & $42.0\%$ & $42.7\%$ & $\mathbf{46.0}\%$ & $45.7\%$ & $\mathbf{51.5}\%$ & $46.2\%$ & $45.2\%$ & $\mathbf{57.0}\%$ & $52.2\%$ & $56.5\%$ \\
& GLM‑4V+ & $37.7\%$ & $38.2\%$ & $\mathbf{45.0}\%$ & $39.8\%$ & $\mathbf{46.5}\%$ & $42.8\%$ & $41.5\%$ & $53.0\%$ & $45.7\%$ & $\mathbf{53.3}\%$ \\
& O1‑Mini & $33.7\%$ & $33.3\%$ & $\mathbf{37.8}\%$ & $33.2\%$ & $\mathbf{38.5}\%$ & $36.0\%$ & $35.5\%$ & $43.5\%$ & $39.3\%$ & $\mathbf{45.2}\%$ \\
& GPT-4o    & $\mathbf{43.0}\%$ & $42.7\%$ & $34.5\%$ & $\mathbf{43.0}\%$ & $36.2\%$ & $39.0\%$ & $37.8\%$ & $49.2\%$ & $36.5\%$ & $\mathbf{52.2}\%$ \\
\bottomrule
\end{tabularx}
\label{tab:final_processed_result}
\end{table*}

\subsection{Main Results (RQ1)}
% Table \ref{tab:main_result} records the attack success rates of multi-agent systems under various agent paradigms, large models, and different attack scenarios. This experiment, by controlling variables, effectively analyzes the robustness of multi-agent systems based on different agent paradigms, the anti-interference capabilities of different LLMs in multi-agent systems, and the destructive impact of attacks from different levels on multi-agent systems.

\paragraph{Overall Analysis.} As shown in Table~\ref{tab:final_processed_result}, reasoning-layer attacks consistently yield the highest ASR across all settings. For example, the Chain-of-Thought Injection Attack (CIA) reaches 78.3\% ASR under the ReAct paradigm with Qwen‑7B, approximately 13 points higher than the best communication attack (SBA 65.0\%) and over 17 points higher than the strongest perception attack (CMA 60.8\%). This pattern shows that perturbing internal reasoning traces directly disrupts agent outputs, while perception and communication disturbances primarily affect information transfer or local understanding, which the system can sometimes recover from through collaboration. From the perspective of \textbf{reasoning paradigms}, Reflexion demonstrates the strongest robustness: under the same CIA attack, its ASR drops to 61.7\% (Qwen‑7B), around 16 points lower than ReAct. Plan‑and‑Solve performs moderately, achieving 69.2\% ASR under Qwen‑7B but remaining sensitive to reasoning errors since incorrect plans propagate through the solution stage. ReAct is the most vulnerable, alternating reasoning and acting without explicit validation, so early perturbations are easily amplified. Finally, \textbf{model scalability} further improves overall resistance. Larger models such as o1‑mini and GPT‑4o consistently yield lower ASR across all layers. CIA falls from 78.3\% (Qwen‑7B) to 65.0\% (GPT‑4o) in ReAct paradigms,  indicating that stronger language models offer better robustness against hierarchical attacks. A similar pattern is observed on the EvoChart-QA benchmark~\cite{huang2025evochart}, as shown in Table 1 in Supplementary Material.

% \paragraph{Perception Layer.}
% Cross‑Modal Attack (CMA) yields the highest ASR in 87\% of tasks, showing that jointly perturbing visual and textual modalities most effectively deceives visual‑language alignment. Under ReAct with Qwen‑7B, CMA reaches 60.8\% ASR, 2.3 points higher than VIA and 5.6 points higher than TIA. Visual‑only (VIA) and text‑only (TIA) attacks are relatively less damaging, as errors at this stage can often be mitigated by later communication or reasoning.

% \paragraph{Communication Layer.}
% Structural Blocking Attack (SBA) substantially outperforms message‑level attacks (SCIA, SMPA). Its ASR reaches 71.8\%, 28 points higher than SCIA and 15 points higher than SMPA under the same conditions. Message‑level attacks tend to produce inconsistencies that can be corrected via cross‑validation or rerouting, whereas structural modifications directly disrupt the communication topology and dominate vulnerabilities in this layer.

% \paragraph{Reasoning Layer.}
% Reasoning‑level attacks cause the most severe degradation. Chain‑of‑Thought Injection (CIA) is the strongest, reaching 78.3\% ASR by directly modifying intermediate reasoning steps; once the trace is corrupted, outputs are hard to repair. Tool Spoofing (TSA) is also highly effective, achieving 76.7\% ASR on Qwen‑7B. These results highlight the reasoning stage as the most fragile component of the MM‑MAS hierarchy.

\paragraph{Perception Layer.} Cross‑Modal Attack (CMA) yields the highest ASR in 87\% of the evaluated tasks, confirming that jointly perturbing visual and textual modalities effectively deceives the agents’ visual‑language alignment. Under the ReAct paradigm with Qwen‑7B, CMA achieves an ASR of 60.8\%, which is 2.3\% higher than VIA and 5.6\% higher than TIA. Visual‑only (VIA) and text‑only (TIA) attacks are relatively less damaging, as errors at this stage can often be alleviated through inter‑agent communication or downstream reasoning. Overall, perception attacks mainly affect local comprehension rather than the global decision process.

\paragraph{Communication Layer.} The Structural Blocking Attack (SBA) achieves a much higher attack success rate (ASR) than message‑level attacks (SCIA, SMPA): 71.8 percent, about 28 points above SCIA and 15 above SMPA under the same conditions. Message‑level attacks mainly cause inconsistent agent responses, which can usually be corrected through cross‑validation or message rerouting. In contrast, structural attacks alter the network topology itself. Among them, agent spoofing is unstable because fake agents generate noisy outputs that others can ignore, whereas link blocking enforces direct disconnection between key agents, cutting off access to correct expertise. These results show that while MM‑MAS retains partial robustness through rerouting and validation mechanisms, topological attacks remain the main weakness in the communication layer.

% Structural Blocking Attack (SBA) shows significantly higher ASR performance than message‑level attacks (SCIA, SMPA). Specifically, the ASR of SBA reaches 71.8\%, which is 28\% higher than SCIA and 15\% higher than SMPA under the same experimental conditions. Message‑level attacks often lead to inconsistent responses among agents, yet these inconsistencies can typically be identified and corrected through cross‑validation or by rerouting messages to alternative agents. In contrast, structural modifications disrupt the network topology itself. Among structural attacks, agent spoofing is relatively unstable because fake agents produce random or inconsistent outputs that other agents can bypass, whereas link blocking enforces a direct disconnection between key agents, preventing the system from accessing correct expertise. These quantitative results indicate that while the MM‑MAS retains partial robustness through rerouting and message validation mechanisms, topological attacks remain the dominant vulnerability within the communication layer.

\paragraph{Reasoning Layer.} Reasoning‑level attacks cause the most severe degradation in system performance. Among them, the Chain‑of‑Thought Injection Attack (CIA) is particularly effective, achieving the highest ASR of 78.3\% in our experiments. This superiority arises because CIA directly modifies intermediate reasoning steps rather than indirectly interfering with memory or external tool usage. Once the reasoning trace is corrupted, agent outputs become unreliable and difficult to correct. In addition, the Tool Spoofing Attack (TSA) also demonstrates strong effectiveness, reaching an ASR of 76.7\% on the Qwen‑7B model, outperforming other approaches under comparable settings. These quantitative results highlight that the reasoning stage is indeed the most fragile component of the MM‑MAS hierarchy, and that reasoning‑level manipulations can cause persistent and hard‑to‑mitigate disruptions.

\subsection{Robustness Analysis (RQ2)}

\paragraph{Task‑Level Effects.} To assess how different attack layers influence overall task performance, we evaluated each reasoning paradigm’s task success rate under attacks, together with its baseline performance under no‑attack conditions. In Table~\ref{tab:origin}, the baseline success rates are around 60\%,suggesting that the three paradigms have comparable performance in the no‑attack setting. When attacks are introduced, success rates drop substantially. The decline is most pronounced for the ReAct paradigm under reasoning‑layer attacks, where performance declines by up to 35\%. Perception ‐ and communication‑layer attacks cause moderate reductions (approximately 25–30\%). These findings demonstrate that reasoning‑level perturbations impose the most severe degradation on task‑level stability, establishing the reasoning layer as the most vulnerable component of the multi‑agent system.

\paragraph{Hallucination‑Level Effects.} We further analyze hallucination errors, which stem from internal instability rather than direct adversarial success. They appear across system layers, including misreading inputs in perception, misunderstanding exchanged messages in communication, and fabricating logic in reasoning. For each layer, ASR is computed after excluding hallucination-induced failures. As shown in Table~\ref{tab:robustness_hallucination}, the hallucination rate decreases from about 8\% for Qwen-7B to around 4\% for GPT-4o, indicating that larger models maintain higher internal stability under adversarial conditions. Reflexion exhibits few external errors but more hallucination‑related ones, whereas ReAct shows fewer internal hallucinations but larger performance drops once its reasoning process is disrupted. These results highlight a trade‑off between external robustness and internal stability, both of which jointly determine overall system reliability.

\begin{table}[h]
\caption{\textbf{TSR of Original MAS experiment.} “N.A.” denotes the baseline without any attack.}
\centering
\small
\setlength{\tabcolsep}{8pt}
% \resizebox{0.9\columnwidth}{!}{%
\begin{tabularx}{\columnwidth}{@{}lcccc@{}}
\toprule
Paradigm & Per. & Comm. & Rea. & N.A. \\
\midrule
ReAct & 29.45\% & 27.58\% & 23.55\% & 58.99\% \\
PlanAndSolve & 34.59\% & 31.99\% & 27.58\% & 60.88\% \\
Reflexion & 33.18\% & 31.43\% & 30.64\% & 61.35\% \\
\bottomrule
\end{tabularx}
% }
\label{tab:origin}
\end{table}

\begin{table}[h]
\caption{Paradigm Robustness and Layer Attack Success/Hallucination Rates.}
\small
\setlength{\tabcolsep}{5.3pt}
\centering
\begin{tabularx}{\columnwidth}{@{}lccccc@{}}
\toprule
Paradigm & LLM  & \multicolumn{3}{c}{ASR} & HER  \\
\cmidrule(lr){3-5} 
& & Per. & Comm. & Rea. &  \\
\midrule
& Qwen-7B & 54.6\% & 57.8\% & 64.8\% & 6.8\% \\
& Qwen-32B & 52.9\% & 56.1\% & 63.2\% & 6.2\% \\
ReAct & GLM-4V+ & 51.2\% & 50.2\% & 59.2\% & 5.7\% \\
& O1-Mini & 44.3\% & 42.9\% & 54.9\% & 4.4\% \\
& GPT-4o & 42.0\% & 42.9\% & 51.9\% & 3.5\% \\
      
\midrule
& Qwen-7B & 47.1\% & 52.1\% & 59.0\% & 8.1\% \\
& Qwen-32B & 45.5\% & 50.4\% & 57.4\% & 7.6\% \\
PlanAndSolve& GLM-4V+ & 43.5\% & 46.6\% & 52.8\% & 6.8\% \\
& O1-Mini & 37.0\% & 40.4\% & 45.2\% & 4.5\% \\
& GPT-4o & 35.0\% & 36.3\% & 41.6\% & 4.3\% \\
             
\midrule
& Qwen-7B & 46.0\% & 48.5\% & 54.6\% & 8.5\% \\
& Qwen-32B & 44.4\% & 46.8\% & 53.0\% & 8.0\% \\
Reflexion& GLM-4V+ & 40.1\% & 41.6\% & 47.8\% & 6.8\% \\
& O1-Mini & 34.9\% & 35.7\% & 41.8\% & 3.8\% \\
& GPT-4o & 39.4\% & 39.0\% & 44.5\% & 4.8\% \\
          
\bottomrule
\end{tabularx}
\label{tab:robustness_hallucination}
\end{table}

% \begin{table}[h]
% \caption{Robustness of Each Paradigm}
% \small
% \centering
% \begin{tabularx}{\columnwidth}{@{}lcc@{}}
% \toprule
% Paradigm & LLM & Success Rate \\
% \midrule
% ReAct & gpt-4o-0806 & 50\% \\
% PlanAndSolve & qwen25-vl-32b & 45\% \\
% Reflexion & qwen25vl-7b & 40\% \\
% \bottomrule
% \end{tabularx}
% \end{table}

% \begin{table}[h]
% \caption{Three-Layer Success and Hallucination Rates}
% \centering
% \small
% \begin{tabularx}{\columnwidth}{@{}lccc@{}}
% \toprule
% & Perception Layer & Communication Layer & Reasoning Layer \\
% \midrule
% Success Rate & 50\% & 40\% & 35\% \\
% Hallucination Rate & 10\% & 15\% & 20\% \\
% \bottomrule
% \end{tabularx}
% \end{table}
\begin{figure}[t]
\centering
\includegraphics[width=\columnwidth]{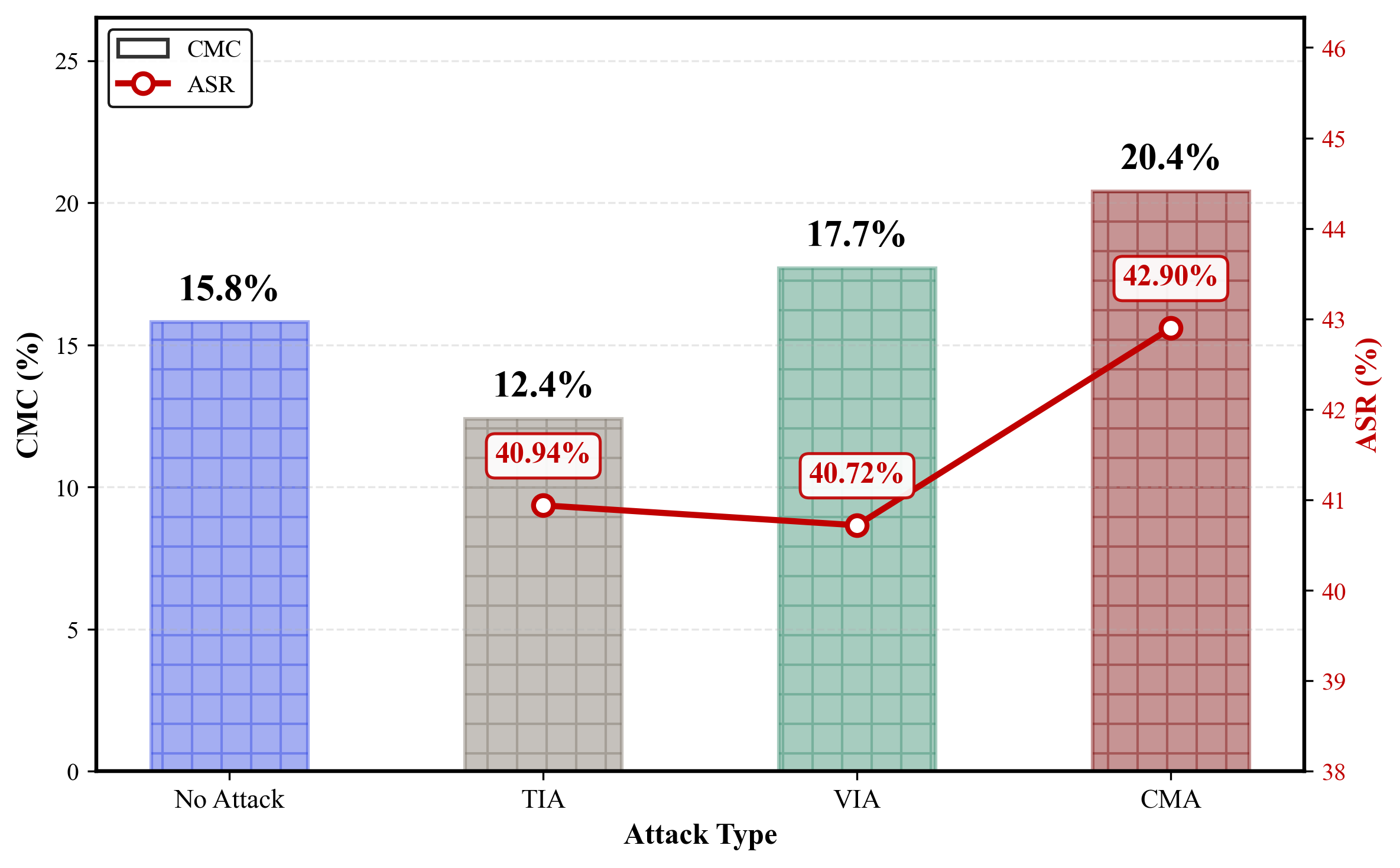}
\caption{Analysis of Multimodal Attack Effects and CMC}
\vspace{-10pt}
\label{fig:modal}
\end{figure}

\subsection{Error Distribution Analysis (RQ3)}
% To analyze failures caused by adversarial attacks, we examined complete multi‑agent interaction traces, including communication, shared context and memory exchange, and used GPT‑4o to identify the underlying error types in each failed case. These attack‑induced failures were categorized into three types: systemic errors (SE) affecting multiple agents collectively, local errors (LE) confined to a single agent, and other errors (OE) representing irregular or feedback‑driven anomalies. According to the results in Table~\ref{tab:error-distribution}, systemic errors are primarily concentrated in the reasoning layer (40.7\%), reflecting the collective dependence of agents on shared reasoning chains. Local errors peak in the communication layer (29.2\%), indicating that inter‑agent messaging disturbances tend to remain localized. Other errors are rare (below 1\%) across all layers, appearing slightly more often in the communication layer, which may reflect occasional feedback amplification during inter‑agent exchanges. Together, these results reveal a dual‑pattern robustness: the system remains partly resilient to local disruptions but strongly vulnerable to reasoning‑level perturbations that synchronize erroneous behavior across agents.

To examine how adversarial perturbations manifest within the multi‑agent system, we classify observed failures into three types. \textbf{Local errors} are mistakes confined to a single agent (e.g., isolated reasoning slips or execution inaccuracies). \textbf{Systemic errors} arise when multiple agents produce the same or mutually consistent wrong outputs, indicating coordinated failures. \textbf{Other errors} denote infrequent or irregular failures such as random fluctuations or occasional feedback‑amplified deviations. 

\begin{figure}[t]
\centering
\includegraphics[width=\columnwidth]{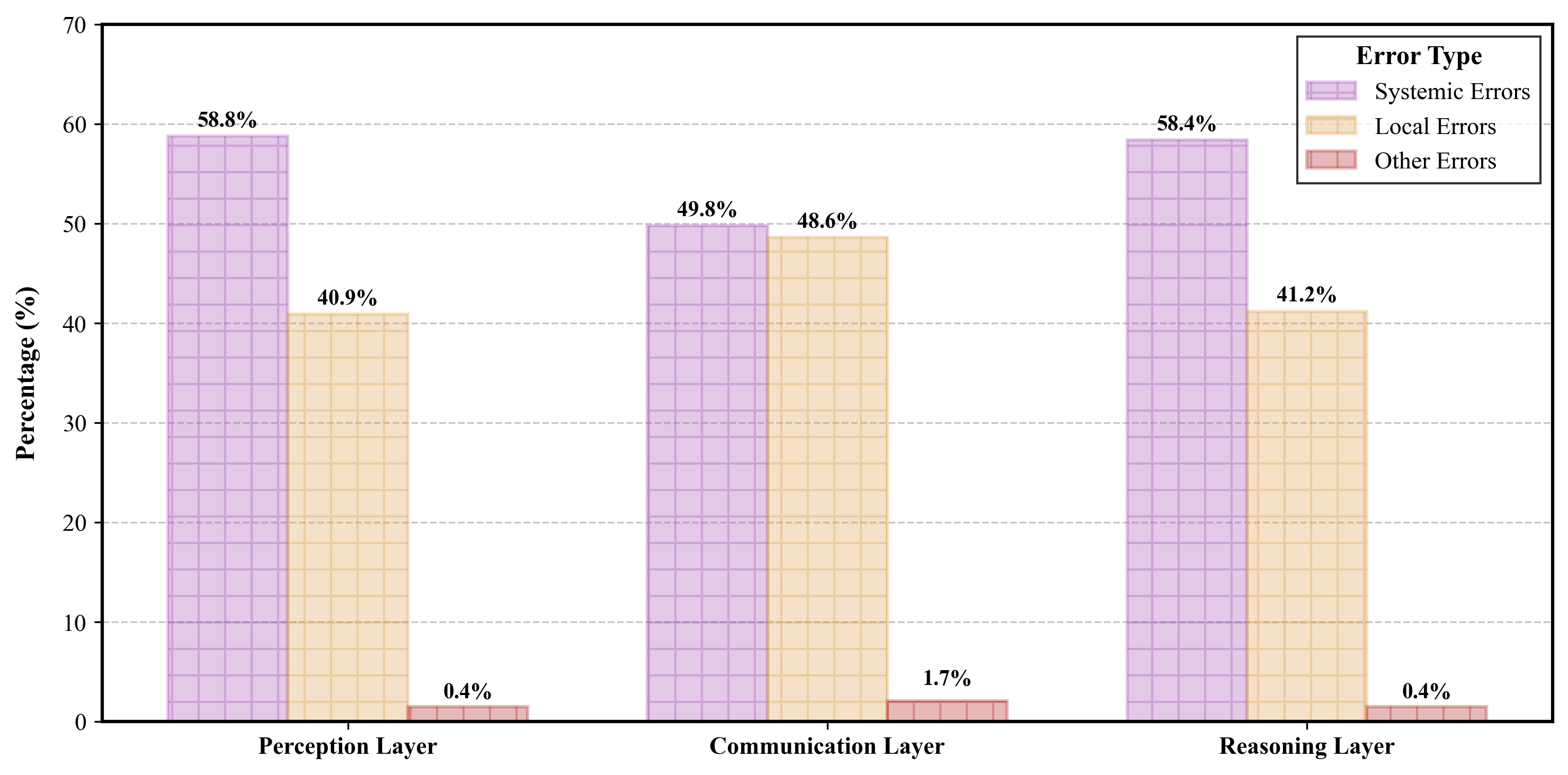}
\caption{Error Distribution Analysis}
\vspace{-10pt}
\label{fig:error-distribution}
\end{figure}

Figure~\ref{fig:error-distribution} summarizes the distribution of these error types across layers. In the perception layer, systemic errors (58.8\%) exceed local errors (40.9\%) by 17.9 points, meaning attacks already trigger coordinated failures at the earliest processing stage. In the communication layer, systemic (49.8\%) and local errors (48.6\%) are nearly balanced, suggesting that message disturbances can either remain localized or propagate depending on interaction patterns. In the reasoning layer, systemic errors (58.4\%) again exceed local ones (41.2\%) by a similar margin (17.2 points), indicating that once reasoning is perturbed, multiple agents tend to reach similar incorrect conclusions. Other errors remain minimal ($<$2\%) in all layers. Overall, systemic errors dominate across layers, showing that our hierarchical attacks consistently disrupt multi‑agent collaboration and that perturbations at any stage can affect global coordination.

\subsection{Cross‑Modal Consistency Analysis (RQ4)}

Figure~\ref{fig:modal} summarizes the effects of perception‑level attacks on cross‑modal consistency (CMC) and attack success rate (ASR). In the benign setting, CMC is 15.8\%, reflecting limited semantic overlap since each query focuses on a local image region. Vision‑only attack (VIA) slightly raises it to 17.7\% (ASR 40.7\%) because visual perturbations steer the agents’ attention toward the same salient area, yielding higher agreement but not correctness. CMA achieves the highest CMC (20.4\%) and ASR (42.9\%), showing that joint textual and visual perturbations generate consistent yet semantically wrong reasoning. Figure~\ref{fig:Perception} further illustrates this effect, where minor image‑text changes lead multiple agents to maintain coherent but false interpretations across modalities.

%% file: sec/5_conclusion.tex
\section{Conclusions}
We proposed HAM\textsuperscript{3}, a hierarchical attack framework designed to identify vulnerabilities in multimodal multi-agent systems across perception, communication, and reasoning layers. Our experiments demonstrate that cross-modal perturbations amplify through fusion, structural communication attacks degrade cooperation, and reasoning-layer interference has the most persistent impact. These findings highlight the cascading nature of vulnerabilities in these systems and provide valuable insights for designing more robust multi-agent intelligence. Additionally, our work reveals how adversarial perturbations at different layers propagate and interact, stressing the need for comprehensive security strategies in multimodal multi-agent systems. This work sets the foundation for future research on improving the robustness of these systems, offering new directions for mitigating emerging vulnerabilities.

% We proposed HAM\textsuperscript{3}, a hierarchical attack framework designed to systematically identify vulnerabilities in multimodal multi‑agent systems across perception, communication, and reasoning layers. Our framework enables the evaluation of both external robustness and internal stability under cross‑modal perturbations. Experiments reveal three key vulnerability patterns: cross‑modal noise is amplified during fusion, structural communication attacks undermine cooperative reasoning, and reasoning‑layer interference exerts the most persistent degradation on task performance. Together, these findings uncover the cascading nature of multimodal vulnerabilities, where minor perturbations at low levels propagate upward and distort collective decision‑making. Our study provides a comprehensive view of system fragility and offers practical insights for developing more resilient collaboration protocols, adaptive message routing, and robust reasoning mechanisms. 